# Learning about individuals from group statistics

**Hendrik Kück** and **Nando de Freitas**
Dept. of Computer Science
University of British Columbia
Vancouver, Canada
{kueck,nando}@cs.ubc.ca

## Abstract

We propose a new problem formulation which is similar to, but more informative than, the binary multiple-instance learning problem. In this setting, we are given groups of instances (described by feature vectors) along with estimates of the fraction of positively-labeled instances per group. The task is to learn an *instance level* classifier from this information. That is, we are trying to estimate the unknown binary labels of individuals from knowledge of group statistics. We propose a principled probabilistic model to solve this problem that accounts for uncertainty in the parameters and in the unknown individual labels. This model is trained with an efficient MCMC algorithm. Its performance is demonstrated on both synthetic and real-world data arising in general object recognition.

## 1 Introduction

Learning models of the relationship between attributes of individuals and the decisions they make is relevant in many contexts. For example, based on factors such as age, education, religion and education it could be of interest to learn the probability of an individual

- voting for the 'purple' party,
- buying 'Brand X' dish-washing detergent,
- moving to another city within the next 5 years.

Having such models allows us to gain an understanding of the motivations behind those decisions, which could for instance help to inform political or marketing strategies.

Unfortunately, it is extremely rare that we have direct information on individual decisions. An expensive solution is to conduct surveys to gather this data, which is not always feasible. In this paper, we propose the use of *indirect* information about the individual decisions, which is often available in the form of group statistics. For instance, we might know that 73% of the people in district $A$ voted 'purple', but only 25% of the inhabitants of district $B$ and 85% in district $C$.

We propose an approach in which *group* statistics in conjunction with pre-existing information about the individuals within the groups are used to learn models relating *individual properties* to *individual decisions*. One example of such pre-existing information is the publicly accessible PUMS (public use microdata sample) data [16] made available by the U.S. Census Bureau, which provides detailed (albeit anonymized) information about individuals in each census tract. It includes data about age, education, income, health and many other continuous and discrete properties of the individuals.

To make our setting more precise, our probabilistic model assumes that individuals can be described by feature vectors. Each individual has an unknown binary label (i.e., the decision in our above example) and, for each group, we are given an estimate of the fraction of individuals with a positive label in the group.

The proposed learning approach is not limited to analyzing human behavior; it also applies to other concept learning tasks arising in artificial intelligence. We provide an example in the context of object class recognition.

The learning problem we are facing in this setting is closely related to the multiple instance (MI) learning problem, which was first introduced by Dietterich et al. in [3]. In the classical MI formulation, binary labels are given for groups of individuals. The binary label of a group is taken to be the disjunction of the unknown individual binary labels. That is, a positive group label indicates that *at least one* of the individuals in the group has a positive label while a negative label implies that *all* individuals in the group have a negative label. Since it was first introduced, this problem has received a lot of attention, and many different algorithms have been proposed for learning in the MI setting (for example [11, 19, 1, 4]).

Several researchers have proposed generalizations of the MI formulation. In [17] and [14] the assumed mapping from individual binary labels to the group label is changed from the logical conjunction used in the classical MI setting to more general threshold functions. A positive group label then indicates that the number of positive instances in the group lies within a certain range. This restriction on the number of positive individual labels in a positive group is similar in spirit to the estimate of the fraction of positive instances in our setting. However, a major difference is that in [17, 14] there is one global threshold or range, whereas we have a different estimate of the ratio of positives for each group and therefor significantly more information.

Most of the published research on MI learning has focused on learning classifiers for *groups*. In [9] a fully probabilistic approach for learning *instance level classifiers* from MI data is presented. Here we adapt the model and learning algorithm presented in [9] to the more informative setting, in which a real value $m \in [0,1]$ indicating the fraction of individuals with a positive label is provided for each group.

## 2 Probabilistic classification

Our goal is to learn a probabilistic model of the relationship between properties of individuals and their binary labels. The conditional probability of interest is $Pr(y=1|\mathbf{x}, \mathcal{D})$, where $\mathbf{x}$ is the feature vector describing an individual, $y$ is the binary label and $\mathcal{D}$ is the given training data. In order to represent this probability distribution, we adopt a parametrized probabilistic model $Pr(y=1|\mathbf{x}, \boldsymbol{\theta})$, where $\boldsymbol{\theta}$ represents the set of model parameters. Instead of finding the one set of parameters $\boldsymbol{\theta}_{ML}$ which best matches the given training data, we take the more principled Bayesian approach and integrate out the uncertainty in the parameters

$$\Pr(y=1|\mathbf{x}, \mathcal{D}) = \int Pr(y=1|\mathbf{x}, \boldsymbol{\theta})\, p(\boldsymbol{\theta}|\mathcal{D})\, d\boldsymbol{\theta}. \quad (1)$$

This marginalization approach is more robust than the maximum likelihood method because the posterior distribution $p(\boldsymbol{\theta}|\mathcal{D})$ is highly multi-modal in our setting.

In Section 3, we first explain our probabilistic model. Section 4 describes the learning algorithm used to compute the posterior distribution $p(\boldsymbol{\theta}|\mathcal{D})$ which then allows us to perform probabilistic classification using Equation (1). In Section 5, we present results on both synthetic and real-world data before we conclude in Section 6.

## 3 Probabilistic model

Our hierarchical probabilistic model is shown in Figure 1. We describe its components subsequently.

The predictive distribution is represented using a real valued function $f$ with parameters $\boldsymbol{\theta} = \{\boldsymbol{\beta}, \boldsymbol{\gamma}\}$ whose output is mapped to a probability as follows:

$$\Pr(y=1|\mathbf{x}, \boldsymbol{\beta}, \boldsymbol{\gamma}) = \Phi(f(\mathbf{x}, \boldsymbol{\beta}, \boldsymbol{\gamma})), \quad (2)$$

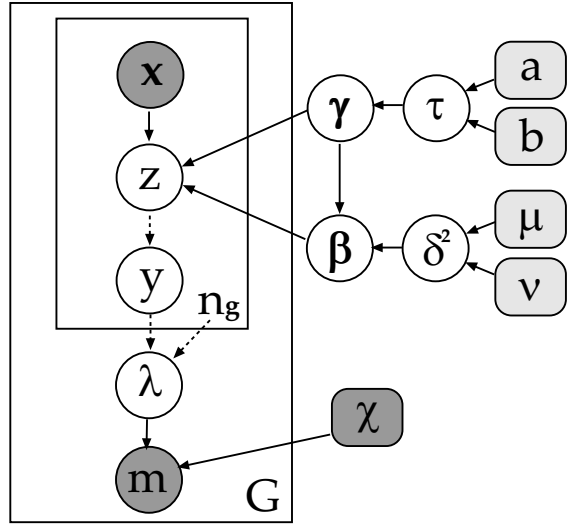

Figure 1: The full graphical model using plate notation. Elements in the outer plate are instantiated for each of the $G$ groups while variables in the inner plate are instantiated for the $n_\mathbf{g}$ individuals in a group. The observations are the individual features $\mathbf{x}$ and the estimate of the ratio of positives per group $m$. $z$ denotes the latent output of a kernel machine with coefficients $\boldsymbol{\beta}$ and kernel selection variables $\boldsymbol{\gamma}$ for input $\mathbf{x}$. This output is mapped to a discrete classification label $y$. The hyper-parameters $\tau$ and $\delta^2$ regularize the model. Finally, $\lambda$ denotes the unknown actual fraction of positives per group. Solid arrows indicate probabilistic dependencies while dashed arrows represent deterministic relationships.

where $\Phi$ is the cumulative distribution function (cdf) of the univariate Normal distribution $\mathcal{N}(0,1)$. $\Phi$ provides a continuous and monotonic mapping from $\mathbb{R}$ to the range $[0,1]$, thus producing a valid probability. The function $\Phi$ in this context is called the *probit link*. Often, the *logistic link* function is used instead, however the probit link is an equally valid choice and its connection to the standard univariate normal distribution will lead to an efficient sampler for our model.

Following Tam, Doucet and Kotagiri [15], the unknown function $f$ is represented with a sparse kernel machine with basis functions centered at the feature vectors $\{\mathbf{x}_1, \ldots, \mathbf{x}_N\}$ of instances from all $G$ groups in the given training data:

$$f(\mathbf{x}, \boldsymbol{\beta}, \boldsymbol{\gamma}) = \sum_{i=1}^{N} \gamma_i \beta_i K(\mathbf{x}, \mathbf{x}_i). \quad (3)$$

Here $K$ is a kernel function, for example a Gaussian kernel $K(\mathbf{x}, \mathbf{x}_i) = e^{-\sigma\|\mathbf{x}-\mathbf{x}_i\|^2}$. $\boldsymbol{\beta} \in \mathbb{R}^N$ is a $N$-dimensional vector of kernel weights and $\boldsymbol{\gamma} \in \{0,1\}^N$ a $N$-dimensional kernel selection vector. For an instance with index $i$ in the training data set, $\gamma_i$ controls whether the kernel basis function located at the corresponding point in feature space is active, and $\beta_i$ controls its weight. In the semi-parametric modeling approach of Equation (3), features are mapped to a high-dimensional nonlinear kernel manifold. Here, there

are as many parameters as data instances. However, the information in the prior and data will force most of the kernel selection variables $\gamma_i$ to be zero and, hence, result in a much lower dimensional manifold.

During training, Equation (2) needs to be evaluated only at the $N$ training data points $\mathbf{X} = \{\mathbf{x}_1, \ldots, \mathbf{x}_N\}$. It is then convenient to express Equation (2) using matrix notation

$$\Pr(y_i = 1|\boldsymbol{\beta}, \boldsymbol{\gamma}) = \Phi(\Psi_{\boldsymbol{\gamma} i} \boldsymbol{\beta}_{\boldsymbol{\gamma}}) \qquad (4)$$

where $\Psi \in \mathbb{R}^{N \times N}$ is the Gram matrix with entries $\Psi_{i,j} = K(x_i, x_j)$. $\Psi_{\boldsymbol{\gamma}}$ is a reduced version of $\Psi$, containing only the columns corresponding to active kernels ($\boldsymbol{\gamma}_i = 1$) and $\Psi_{\boldsymbol{\gamma} j}$ is the $j$-th row of this matrix. $\boldsymbol{\beta}_{\boldsymbol{\gamma}}$ is the reduced version of $\boldsymbol{\beta}$ containing only the coefficients of the active kernels. In the following we never actually use the full vector $\boldsymbol{\beta}$ and for notational simplicity will therefore drop the subscript and from here on use $\boldsymbol{\beta}$ to refer to the reduced vector.

We follow a hierarchical Bayesian strategy, where the unknown parameters $\boldsymbol{\beta}$ and $\boldsymbol{\gamma}$ are drawn from appropriate prior distributions. We place a maximum entropy g-prior [18] on the regression coefficients

$$p(\boldsymbol{\beta}|\boldsymbol{\gamma}, \delta^2) = \mathcal{N}\left(\mathbf{0}, \delta^2(\Psi_{\boldsymbol{\gamma}}^T \Psi_{\boldsymbol{\gamma}})^{-1}\right) \qquad (5)$$

and assign an inverse Gamma prior to the regularization parameter $\delta^2$

$$p(\delta^2) = \mathcal{IG}\left(\frac{\mu}{2}, \frac{\nu}{2}\right) \qquad (6)$$

with fixed hyper-parameters $\mu$, $\nu$ typically set to near-uninformative values (for example $\mu = \nu = 1$). Each $\gamma_i$ follows a Bernoulli distribution with success rate $\tau \in [0, 1]$, which in turn follows a Beta distribution with parameters $a, b \geq 1$. This allows the model to adapt to the data while giving the user some control over the desired fraction of active kernels. By integrating out $\tau$ we get:

$$p(\boldsymbol{\gamma}) = \int p(\boldsymbol{\gamma}|\tau)p(\tau)d\tau = \frac{\Gamma(k+a)\,\Gamma(N-k+b)}{\Gamma(N+a+b)}, \qquad (7)$$

where $k = \sum_i \gamma_i$ is the number of active kernels, i.e. the number of non zero elements in $\boldsymbol{\gamma}$.

This hierarchical Bayesian approach allows us to be more vague in the specification of our prior beliefs. Instead of choosing a fixed value $\tau$ for the percentage of kernels that we believe should be active in the model, we can choose a diffuse distribution over $\tau$ (using small values for $a$ and $b$) instead. This encodes our lack of a priori knowledge of $\tau$ and lets the data automatically determine the complexity of the model according to the principle of Occam's razor.

To facilitate efficient computation, we employ the data augmentation trick first introduced by Nobel laureate Daniel McFadden [12], which in the context of this specific model was also used in [15]. The probabilistic model discussed so far is augmented by introducing the continuous latent variable $z \in \mathbb{R}$, which can be seen as a continuous version of the binary label $y$ satisfying $y = sign(z)$. It then follows, from the choice of the probit link in Equation (2), that in order to keep the marginal distribution of the other variables invariant, $z$ has to be distributed as

$$z \sim \mathcal{N}(f(\mathbf{x}, \boldsymbol{\beta}, \boldsymbol{\gamma}), 1)$$

The joint distribution for the full set of augmentation variables $\mathbf{Z} = \{z_1, \ldots, z_N\}$ for the $N$ instances in the training data is then

$$p(\mathbf{Z}|\boldsymbol{\beta}, \boldsymbol{\gamma}, \mathbf{X}) = \mathcal{N}(\Psi_{\boldsymbol{\gamma}} \boldsymbol{\beta}_{\boldsymbol{\gamma}}, \mathbf{I}_N), \qquad (8)$$

where $\mathbf{I}_N$ is the $N$-dimensional identity matrix. It will be shown in Section 4 that conditioning on $\mathbf{Z}$ makes the posterior of the high-dimensional coefficients $\boldsymbol{\beta}$ a Gaussian distribution that can be obtained analytically. Hence this augmentation strategy replaces the problem of sampling from a high dimensional, highly correlated, distribution with the much simpler problem of sampling lower dimensional variables.

### 3.1 Bringing in the evidence

The model discussed up to this point is nearly identical with the one presented in [15], where it is used in the supervised classification context, in which the individual labels $\mathbf{Y} = \{y_1, \ldots, y_N\}$ are given for the training instances. In [9] we adapted the model to the multiple instance setting. Here, instead of a binary label, we are given $m_j \in [0, 1]$, an estimate of the fraction of positive instances for each group in the training data. Additionally, we have a parameter $\chi$ quantifying the confidence in these guesses. Higher values indicate higher confidence, while $\chi = 0$ indicates a complete lack of confidence. Our training data is thus given by

$$\mathcal{D} = \{\mathbf{X}, M, \chi\} = \{\mathbf{X}_{\mathbf{g}_1}, \ldots, \mathbf{X}_{\mathbf{g}_G}, m_1, \ldots, m_G, \chi\},$$

where $\mathbf{X}_{\mathbf{g}_j}$ is the set of feature vectors describing the instances in group $j$. Note that while we use a global confidence parameter $\chi$ here, it is straightforward to modify our model and training algorithm to deal with separate confidence estimates for each group, should they be available. The given estimate $m_j$ for a group $j$ is modeled as a noisy measurement of the unknown actual ratio of positives (denoted $\lambda_j$) in that group. This models measurement errors (such as miscounts of election votes) as well as other sources of uncertainty. The value $\lambda_j$ is deterministically computed from $\mathbf{Z}_{\mathbf{g}_j}$, the set of augmentation variables for the instances in group $j$

$$\lambda_j = \frac{1}{n_{\mathbf{g}j}} \sum_{i \in \mathbf{g}_j} \mathbb{I}_{(0,\infty)}(z_i), \qquad (9)$$

where $n_{\mathbf{g}j}$ is the number of instances in group $j$. Note that we implicitly integrated out $y$ in Equation 9. The Beta distribution was chosen to model the measurement process producing the estimate $m_j$. Its two parameters are deterministically computed from $\chi$ and $\lambda_j$ such that the mode of the distribution is located at $\lambda_j$ while $\chi$ controls the peakedness.

$$p(m_j|\lambda_j, \chi) = \mathcal{B}(\chi\lambda_j + 1, \chi(1-\lambda_j) + 1)$$
$$\propto m_j^{\chi\lambda_j}(1 - m_j)^{\chi(1-\lambda_j)}.$$

## 4 Computation

In order to perform probabilistic inference on individuals, we need a way to evaluate Equation (1). However, for the model discussed in the previous section, the integral involved turns out to be intractable.

Instead, we use Monte Carlo simulation to generate samples $\{\boldsymbol{\theta}^{(1)}, \ldots, \boldsymbol{\theta}^{(T)}\}$ from the posterior distribution of the model parameters

$$\boldsymbol{\theta}^{(t)} \sim p(\boldsymbol{\theta}|\mathcal{D})$$

which allows us to approximate Equation (1) as

$$\Pr(y = 1|\mathbf{x}, \mathcal{D}) \approx \frac{1}{T} \sum_{t=1}^{T} \Pr(y = 1|\mathbf{x}, \boldsymbol{\theta}^{(t)}). \quad (10)$$

This approximation converges to the true solution by the Strong Law of Large Numbers.

Although we are only interested in $\boldsymbol{\theta}^{(t)} = \{\boldsymbol{\beta}^{(t)}, \boldsymbol{\gamma}^{(t)}\}$, integrating out all other latent variables turns out to be intractable. Instead, we sample from the full joint posterior $p(\boldsymbol{\beta}, \boldsymbol{\gamma}, \mathbf{Z}, \delta^2|\mathcal{D})$ and then marginalize $\mathbf{Z}$ and $\delta^2$ by simply ignoring these components in the generated samples. Note that $\tau$ was already integrated out in Equation (7) and both $\lambda$ and $y$ follow deterministically from $\mathbf{Z}$ and thus do not need to be sampled.

Gibbs sampling [6] is a well known MCMC technique in which the individual variables are sampled in turn from their full conditional distributions. As in [15], we are using a *blocked* Gibbs sampler, in which $\{\boldsymbol{\gamma}, \boldsymbol{\beta}\}$ are sampled together as one block and $\{\delta^2, \mathbf{Z}\}$ as a second block. Sampling variables jointly in blocks results in a sampler with much better mixing probabilities by reducing the correlation amongst samples [10, 7].

The joint conditional distribution for $\{\boldsymbol{\gamma}, \boldsymbol{\beta}\}$ factors as

$$p(\boldsymbol{\beta}, \boldsymbol{\gamma}|\delta^2, \mathbf{Z}, \mathbf{X}) = p(\boldsymbol{\beta}|\boldsymbol{\gamma}, \delta^2, \mathbf{Z}, \mathbf{X})p(\boldsymbol{\gamma}|\delta^2, \mathbf{Z}, \mathbf{X}).$$

It should be noted by looking at the graphical model in Figure 1 that $\delta^2$ and $\mathbf{Z}$ are conditionally independent, given $\boldsymbol{\beta}$ and $\boldsymbol{\gamma}$. We use this fact when sampling $\{\delta^2, \mathbf{Z}\}$ as

$$p(\delta^2, \mathbf{Z}|\boldsymbol{\beta}, \boldsymbol{\gamma}, \mathbf{X}, m) = p(\delta^2|\boldsymbol{\beta}, \boldsymbol{\gamma})p(\mathbf{Z}|\boldsymbol{\gamma}, \boldsymbol{\beta}, \mathbf{X}, \mathbf{M}).$$

```
1  initialize γ⁽⁰⁾, δ²⁽⁰⁾, Z⁽⁰⁾
2  for t = 1 to T:
3      Sample:
4          γ⁽ᵗ⁾ ~ p(γ|δ²⁽ᵗ⁻¹⁾, Z⁽ᵗ⁻¹⁾, X)
5          β⁽ᵗ⁾ ~ p(β|γ⁽ᵗ⁾, δ²⁽ᵗ⁻¹⁾, Z⁽ᵗ⁻¹⁾, X)
6          δ²⁽ᵗ⁾ ~ p(δ²|β⁽ᵗ⁾, γ⁽ᵗ⁾)
7          for j = 1 to G:
8              Z_gⱼ⁽ᵗ⁾ ~ p(Z_gⱼ|γ⁽ᵗ⁾, β⁽ᵗ⁾, X_gⱼ, mⱼ)
```
Listing 1: Blocked Gibbs sampler for sampling from the joint posterior distribution $p(\boldsymbol{\beta}, \boldsymbol{\gamma}, \mathbf{Z}, \delta^2|\mathcal{D})$.

While the observation $m_j$ introduces conditional dependencies amongst the augmentation variables $\mathbf{Z}_{\mathbf{g}_j}$ in group $j$, the augmentation variables of different groups are independent

$$p(\mathbf{Z}|\boldsymbol{\gamma}, \boldsymbol{\beta}, \mathbf{X}, \mathbf{M}) = \prod_{j=1}^{G} p(\mathbf{Z}_{\mathbf{g}_j}|\boldsymbol{\gamma}, \boldsymbol{\beta}, \mathbf{X}_{\mathbf{g}_j}, m_j).$$

The overall blocked Gibbs sampler is given in Listing 1. In the following, we provide the conditional distributions involved (lines 4 to 8) and outline the techniques for sampling from them.

Sampling the $N$-dimensional binary kernel selection vector $\boldsymbol{\gamma}$ is the most involved and computational intensive part of the overall Gibbs sampler. Its conditional distribution is

$$p(\boldsymbol{\gamma}|\delta^2, \mathbf{Z}, \mathbf{X}) \propto (1 + \delta^2)^{-\frac{K}{2}} e^{\frac{1}{2}\left(\frac{\delta^2}{1+\delta^2}\mathbf{Z}^T \Psi_{\boldsymbol{\gamma}}(\Psi_{\boldsymbol{\gamma}}^T\Psi_{\boldsymbol{\gamma}})^{-1}\Psi_{\boldsymbol{\gamma}}^T\mathbf{Z}^T\right)}$$
$$\times \frac{\Gamma(K+a)\,\Gamma(N-K+b)}{\Gamma(N+a+b)}. \quad (11)$$

We use an efficient Metropolis Hastings within Gibbs sampler very similar to the one described in [15] for sampling from this distribution. A detailed description of this sampler and the full derivation of Equation (11) can be found in [8].

The full conditional distribution for the kernel weights $\boldsymbol{\beta}$ follows from Bayes rule:

$$p(\boldsymbol{\beta}|\boldsymbol{\gamma}, \delta^2, \mathbf{Z}, \mathbf{X}) \propto p(\mathbf{Z}|\boldsymbol{\beta}, \boldsymbol{\gamma}, \mathbf{X})p(\boldsymbol{\beta}|\boldsymbol{\gamma}, \delta^2).$$

Thanks to the data augmentation trick, both the likelihood and prior (Equations (8) and (5)) are Normal distributions. Therefore the posterior can be computed analytically and efficiently sampled from, as

$$\boldsymbol{\beta} \sim \mathcal{N}\left(\frac{\delta^2}{1+\delta^2}(\Psi_{\boldsymbol{\gamma}}^T\Psi_{\boldsymbol{\gamma}})^{-1}\Psi_{\boldsymbol{\gamma}}^T\mathbf{Z}, \frac{\delta^2}{1+\delta^2}(\Psi_{\boldsymbol{\gamma}}^T\Psi_{\boldsymbol{\gamma}})^{-1}\right)$$

Similarly, for the conditional distribution of the regularization parameter $\delta^2$ we have

$$p(\delta^2|\boldsymbol{\beta}, \boldsymbol{\gamma}) \propto p(\boldsymbol{\beta}|\boldsymbol{\gamma}, \delta^2)p(\delta^2).$$

Since the inverse gamma distribution of $p(\delta^2)$ is conjugate to the Normal distribution of $p(\boldsymbol{\beta}|\boldsymbol{\gamma}, \delta^2)$ (Equations (5) and (6)), the posterior can be computed analytically and sampled from using a standard inverse gamma sampler

$$\delta^2 \sim \mathcal{IG}\left(\frac{1}{2}(\mu + K + 1), \frac{1}{2}(\nu + \|\Psi_{\boldsymbol{\gamma}}\boldsymbol{\beta}\|^2)\right).$$

The joint conditional distribution of the augmentation variables $\mathbf{Z}_{\mathbf{g}_j}$ for the instances in group $j$ has the form

$$p(\mathbf{Z}_{\mathbf{g}_j}|m_j, \mathbf{X}_{\mathbf{g}_j}, \boldsymbol{\beta}, \boldsymbol{\gamma}) \propto p(m_j|\mathbf{Z}_{\mathbf{g}_j})p(\mathbf{Z}_{\mathbf{g}_j}|\mathbf{X}_{\mathbf{g}_j}, \boldsymbol{\beta}, \boldsymbol{\gamma})$$
$$= \mathcal{B}(\chi\lambda_j + 1, \chi(1-\lambda_j) + 1)$$
$$\times \prod_{i \in \mathbf{g}_j} p(z_i|\mathbf{x}_i, \boldsymbol{\beta}, \boldsymbol{\gamma}),$$

where $\lambda_j$ is computed from $\mathbf{Z}_{\mathbf{g}_j}$ using Equation (9). This posterior distribution is a $n_{\mathbf{g}}$-dimensional multivariate Gaussian, which is scaled by different constants in different orthants of the space. The orthants which correspond to the guessed fraction of positives, $m_j$, will have larger scaling factors than those that do not. Since the prior and likelihood do not combine, we cannot directly sample from this distribution. Instead we use a Metropolis within Gibbs sampler with an isotropic Normal distribution as proposal. That is, a new set of candidate values $\mathbf{Z}'_{\mathbf{g}_j}$ is generated based on the previous sample $\mathbf{Z}^{(t)}_{\mathbf{g}_j}$ using

$$\mathbf{Z}'_{\mathbf{g}_j} \sim q(\mathbf{Z}^{(t)}_{\mathbf{g}_j}, \mathbf{Z}'_{\mathbf{g}_j}) = \mathcal{N}(\mathbf{Z}^{(t)}_{\mathbf{g}_j}, c^2 \mathbf{I}),$$

where $c$ controls the variance of the proposal. The proposal is accepted with acceptance rate

$$A = min\left(1, \frac{p(\mathbf{Z}'_{\mathbf{g}_j}|m_j, \mathbf{X}_{\mathbf{g}_j}, \boldsymbol{\beta}, \boldsymbol{\gamma})}{p(\mathbf{Z}^{(t)}_{\mathbf{g}_j}|m_j, \mathbf{X}_{\mathbf{g}_j}, \boldsymbol{\beta}, \boldsymbol{\gamma})}\right)$$

We use a value of $c = 2.4\, n_{\mathbf{g}_j}^{-\frac{1}{2}}$ for the proposal variance, which was shown to be optimal for sampling from the unit variance multivariate normal distribution [5]. While this only implies optimality for the limiting case of $\chi = 0$, this proposal distribution does result in reasonable acceptance rates in practice.

In the special cases $m_j = 1$ and $m_j = 0$ it follows from the properties of the Beta distribution that $\lambda_j = m_j$. As a consequence, all instances in the group are enforced to be positive (resp. negative) for $m_j = 1$ (resp. $m_j = 0$) and their augmentation variables $z_i$ are sampled independently from truncated univariate Normal distributions as described in [9]. This provides a straightforward way to incorporate any supervised data, should it be available.

## 5 Experiments

We demonstrate the performance of our proposed probabilistic model and training algorithm on both synthetic and real-world data. In all these experiments we actually know the binary labels of the individual instances and use them for evaluating the learned probabilistic classifiers. We would like to stress however, that our learning algorithm does *not* have access to these labels but is only given the fraction of positives per group (or an estimate thereof).

### 5.1 Synthetic data

In our first experiment we tested our proposed approach on the simple synthetic dataset shown in Figure 2(a). In this example, the individuals are described by 2D feature vectors. The dataset consists of only 3 groups, the statistics of which are given in Table 1.

|  | A | B | C |
|---|---|---|---|
| No. of instances $n_{\mathbf{g}_j}$ | 19 | 16 | 20 |
| Positive fraction $m_j$ | 0.73 | 0.25 | 0.85 |

Table 1: Group statistics for the dataset in Figure 2(a).

Note that each of the groups contains a mixture of positive and negative instances, which is typical for many practical applications. In an election, for example, it is extremely unlikely that everybody votes identically in a district. In the multiple instance framework, each group would hence have a positive label and this MI data thus would thus be uninformative when trying to infer the labels of individuals.

Using the MCMC algorithm described in Section 4, we generated 1000 samples from $\Pr(\boldsymbol{\theta}|\mathcal{D})$ after a burn in period of 1000 simulation steps. We chose a Gaussian kernel, uninformative hyper-parameter values $a = b = \mu = \nu = 1.0$ and a confidence value of $\chi = 1000$. The run time for the simulation was 6 seconds on a 2.6 Ghz Pentium 4. Figure 2(b) shows the learned predictive distribution $\Pr(y = 1|\mathbf{x}, \mathcal{D})$ evaluated using Equation (10) and the generated samples. The distribution correctly assigns high probability to regions of the feature space containing predominantly positive instances. The flexibility of our semi-parametric model allows to recover the non-linear decision manifold in this example. At the same time, the hierarchical priors regularize the solution and prevent excessive complexity. The computed solution is fairly sparse with on average 15 active kernels.

In a second experiment shown in Figure 3 we generated a synthetic 2D dataset, in which we actually have a large number of groups with exclusively negative instances and a small number of groups with both positive and negative instances. In this example, the positive and negative instances are not nicely separated in the feature space. Instead, a few positive instances are embedded in a large diffuse mass of negatives as shown in Figure 3(a). This property is typical of many real-world concept learning problems in the multiple instance setting. Figure 3(b) visualizes the semi-supervised information provided by the bi-

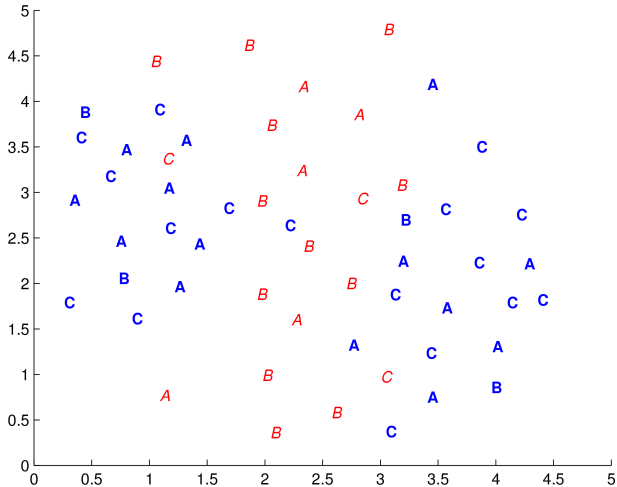 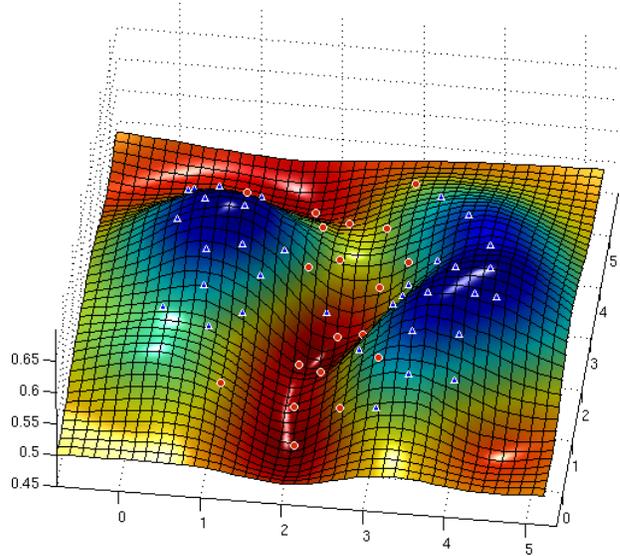

(a) Synthetic dataset consisting of 2D feature vectors belonging to 3 groups ($A$, $B$ and $C$). Each instance has a positive (bold blue letters) or negative (red italicized) label.

(b) The predictive distribution $\Pr(y = 1|\mathbf{x}, \mathcal{D})$ learned by our approach for the data set shown in (a). The location of the positive (blue triangles) and negative (red circles) instances are shown superimposed.

Figure 2: Experiment on a small synthetic dataset with 3 groups of 2D instances. Our learning algorithm learned the predictive distribution shown in (b) from the location of the instances in the 3 groups shown in (a) and the group statistics given in Table 1. *Without knowledge of the instance labels* it managed to reconstruct a predictive distribution which correctly assigns high probability to regions of the feature space containing predominantly positive instances.

nary group labels. The predictive distribution learned from this multiple instance data using the approach described in [9] with a Gaussian kernel is shown in Figure 3(c). Figure 3(d) on the other hand shows the distribution learned by the approach described in this paper (with the same Gaussian kernel) when additionally the fractions of positives $m_j$ were provided for all groups. Clearly, the additional information helped to learn a much improved predictive distribution in which the cluster of positive instances is nicely reconstructed.

### 5.2 Learning object recognition models from annotated images

Finally, we explore the performance of the proposed approach when applied to the task of learning models for object class recognition from images with annotations. Our training data in this experiment consists of 200 annotated images from the Corel database. Each image is segmented into on average about 10 image regions using NCuts [13] and a feature vector describing each region is computed (for more details see [8, 9]). The goal is to learn a probabilistic *image region classifier* for one annotation word/object class at a time. Consider for example the object class 'fox'. Such a classifier allows us to label individual image regions as 'fox' or 'not fox,' making it possible to not only detect but also locate a fox in an image. Learning such a classifier from annotated images constitutes a multiple instance learning problem, in which images are treated as groups of image regions. The binary labels of the image regions (does an image region show a fox or not?) are generally unknown, but the binary labels of whole images are provided by the annotations. This learning problem is similar to the synthetic dataset in Figure 3 in that positive and negative instances can not be expected to be neatly separated in the feature space. Nevertheless, the probabilistic MI learning approach proposed in [9], managed to learn classifiers from this data, which performed quite well.

Even though in this setting we do not know which fraction of the image regions in a given 'fox' image actually show a fox, we can make an educated guess based on the number of image regions and the number of words in the image's annotation: We work from the simplifying assumption that the regions in an image are equally distributed amongst the image's annotation words. That is, for an image where the word 'fox' appears in the annotation as one of $w$ words, we choose the estimate of the fraction of positive image regions to be $m_j = \frac{1}{w}$. Of course this only provides us with a crude estimate of the real fraction of positives. As the results in Figure 4 demonstrate, using these guessed fractions of positives in the approach presented in this paper resulted in improved classification performance for most

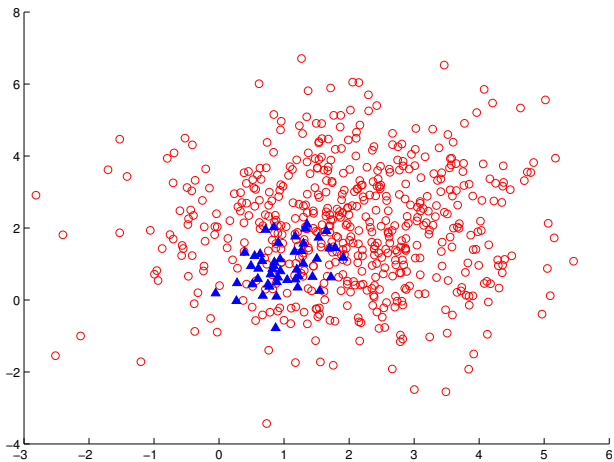

(a) Labeled instances in a synthetic dataset with 600 2D instances. Positive (blue triangles) and negative instances (red circles) were generated from 2 Gaussians. The two classes are not nicely separated in feature space but a cluster of positive instances is embedded in a large diffuse mass of negatives.

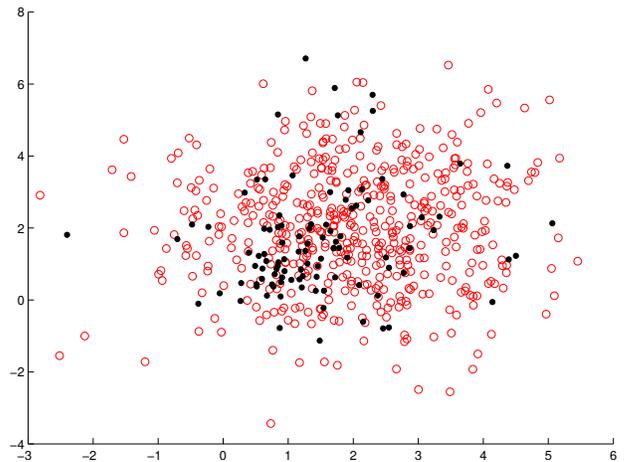

(b) The instances are arranged into 75 groups of 8 instances each. This figure shows the semi-supervised information provided by the binary group labels in the MI setting. For the 62 negative groups, all instances are known to be negative (red circles). The instances in the 13 positive groups can be either positive or negative (black dots).

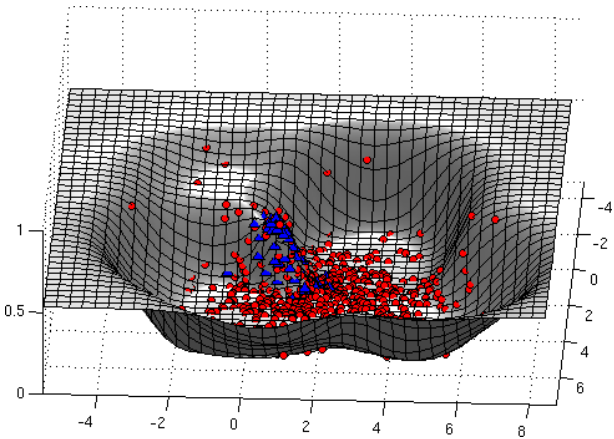

(c) The predictive distribution learned from binary group labels using the MI learning algorithm proposed in [9]. The locations of the labeled training data instances are superimposed. The learned distribution does not quite manage to reconstruct the positive cluster which indicates that the binary group labels do not provide sufficient information about the individual instance labels.

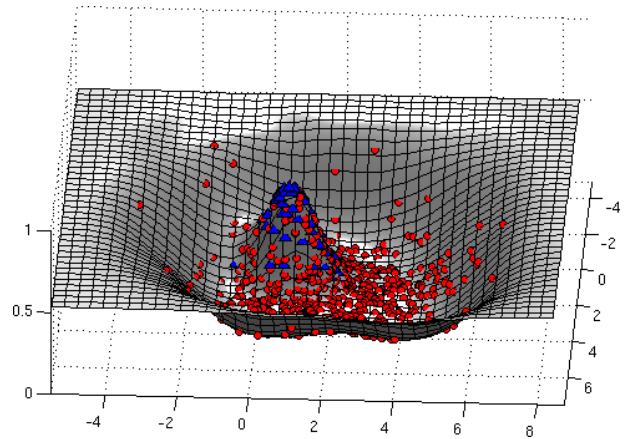

(d) The predictive distribution learned using the approach described in this paper when the fractions of positives $m_j$ for each group are given. A confidence value of $\chi = 1000$ was used. The additional information clearly helps to reconstruct a better predictive distribution in this case. The positive cluster is nicely separated from the surrounding negative instances.

Figure 3: Experiment on a synthetic 2D data set demonstrating the benefit of using information about the fractions of positives as compared to binary group labels. In the synthetic data set used here most of the groups contain exclusively negative instances (Figures (a)(b)). The binary group labels given in multiple instance learning do not provide sufficient information about the individual instance labels in this case, as can be seen in (c). If, on the other hand, for each group, the fraction of positive instances in the group is known, a good probabilistic classifier can be learned from this information using the approach described in this paper as shown in (d). Measuring the classification performance in terms of AUC (area under the receiver operator curve) yields a value of 0.922 for the probabilistic classifier visualized in (d) compared to 0.885 for the one in (c), indicating significantly better classification performance. AUC is a widely recognized measure for comparing the performance of probabilistic classifiers independent of a fixed decision thresholds [2]. A value of 1 indicates perfect classification while 0.5 corresponds to random guessing.

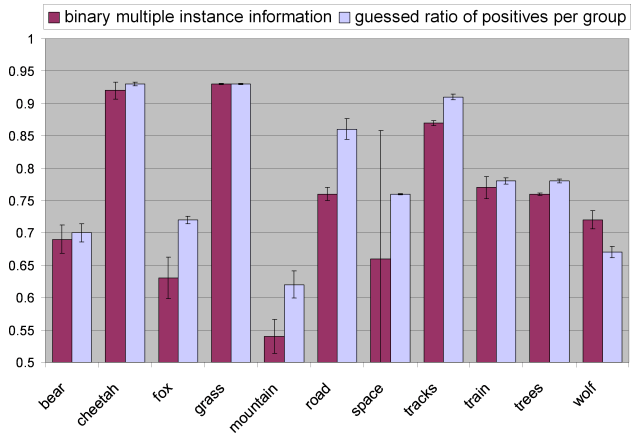

Figure 4: Probabilistic image region classifiers were learned for different object classes/annotation words from a set of 200 annotated images with in total 2070 image regions. The chart compares the performance of classifiers learned from binary group labels using the approach described in [9] with classifiers learned from guessed fractions of positive regions per image using the approach presented in this paper. The classification performance in terms of AUC (area under the ROC curve) is evaluated on a test set of about 1000 manually labeled image regions. The columns display the mean AUC value averaged across 10 runs while the error bars show 1 standard deviation. Although the fractions of positive instances are estimated based on a rather crude assumption and thus tend to not be very accurate, they still lead to improved probabilistic classifiers for most object classes.

object categories when compared to the classifiers learned from MI data. The additional information also seems to have the effect of reducing the variance across runs. Both learning algorithms in this comparison were run with the same hyper-parameters and a sigmoidal kernel. We used a confidence parameter $\chi = 1000$ in our proposed model. 10 000 samples were collected after a 10 000 step burn in period which took between 5-10 minutes (depending on the object category) with both approaches.

## 6  Conclusion

We have introduced a novel problem formulation with relevant real-world applications in social sciences, demographic analysis and marketing research as well as cognitive vision. In this setting we take advantage of group statistics to infer information about individuals.

We developed a principled probabilistic model which fully accounts for the uncertainty in the binary labels of individuals. It has significant modeling power due to a semi-parametric representation while achieving sparsity using a hierarchical prior. An efficient MCMC sampler for training this model was presented.

The problem formulation we introduce can be seen as a more informative variant of the multiple instance learning problem, and we verify on synthetic and real world data that our learning approach effectively leverages this advantage into better probabilistic classifiers.

## References

[1] S. Andrews and T. Hofmann. Multiple instance learning via disjunctive programming boosting. In *Advances in Neural Information Processing Systems (NIPS*16)*, 2004.

[2] A. Bradley. The use of the area under the ROC curve in the evaluation of machine learning algorithms. *Pattern Recognition*, 30:1145–1159, 1997.

[3] T. G. Dietterich, R. H. Lathrop, and T. Lozano-Perez. Solving the multiple instance learning with axis-parallel rectangles. *Artificial Intelligence*, 89(1/2):31–71, 1997.

[4] T. Gartner, P. A. Flach, A. Kowalczyk, and A. J. Smola. Multi-instance kernels. In *19th International Conference on Machine Learning*, pages 179–186, 2002.

[5] A. Gelman, J. B. Carlin, H. S. Stern, and D. B. Rubin. *Bayesian Data Analysis*. Chapman & Hall/CRC, 1995.

[6] S. Geman and D. Geman. Stochastic relaxation, Gibbs distributions, and the Bayesian restoration of images. *IEEE Trans. Pattern Anal. Machine Intell.*, 6(6):721–741, 1984.

[7] F. Hamze and N. de Freitas. From fields to trees. In *Proceedings of the 20th Annual Conference on Uncertainty in Artificial Intelligence (UAI-04)*, pages 243–250, Arlington, Virginia, 2004. AUAI Press.

[8] H. Kück. Bayesian formulations of multiple instance learning with applications to general object recognition. Master's thesis, University of British Columbia, 2004.

[9] H. Kück, P. Carbonetto, and N. de Freitas. A constrained semi-supervised learning approach to data association. In *European Conference for Computer Vision (ECCV)*, 2004.

[10] J. Liu, W. H. Wong, and A. Kong. Covariance structure of the Gibbs sampler with applications to the comparisons of estimators and augmentation schemes. *Biometrika*, 81(1):27–40, 1994.

[11] O. Maron and A. L. Ratan. Multiple-instance learning for natural scene classification. In *Proc. 15th International Conf. on Machine Learning*, pages 341–349. Morgan Kaufmann, San Francisco, CA, 1998.

[12] D. McFadden. A method of simulated moments for estimation of discrete response models without numerical integration. *Econometrica*, 57:995–1026, 1989.

[13] J. Shi and J. Malik. Normalized cuts and image segmentation. In *IEEE Conference on Computer Vision and Pattern Recognition*, pages 731–737, 1997.

[14] Q. Tao, S. Scott, N. V. Vinodchandran, and T. T. Osugi. Svm-based generalized multiple-instance learning via approximate box counting. In *ICML '04: Twenty-first international conference on Machine learning*. ACM Press, 2004.

[15] S. S. Tham, A. Doucet, and K. Ramamohanarao. Sparse Bayesian learning for regression and classification using Markov chain Monte Carlo. In *International Conference on Machine Learning*, pages 634–641, 2002.

[16] U.S. Census Bureau Public Information Office. Public use microdata sample (PUMS), 2000.

[17] N. Weidmann, E. Frank, and B. Pfahringer. A two-level learning method for generalized multi-instance problems. *Lecture Notes in Computer Science*, 2837:468 – 479, 2003.

[18] A. Zellner. On assessing prior distributions and bayesian regression analysis with g-prior distributions. In *Bayesian Inference and Decision Techniques: Essays in Honor of Bruno de Finetti*, pages 233–243. P. K. Goel and A. Zellner (eds), 1986.

[19] Q. Zhang and S. Goldman. EM-DD: An improved multiple-instance learning technique. In *Neural Information Processing Systems 14*, 2001.